\documentclass[10pt,conference]{IEEEtran}

\usepackage{amsmath}
\usepackage{amssymb}
\usepackage{graphicx}
\usepackage{booktabs}
\usepackage{multirow}
\usepackage{makecell}
\usepackage{array}
\usepackage{tabularx}
\usepackage{float}
\graphicspath{{res/}{media/}{figures/}}
\newcolumntype{L}[1]{>{\raggedright\arraybackslash}p{#1}}
\newcolumntype{P}[1]{>{\raggedright\arraybackslash}p{#1}}
\usepackage{url}
\usepackage[hidelinks]{hyperref}
\begin{document}

\title{How Robust Are CNNs and Vision Transformers for Alzheimer's MRI Classification Across Datasets?}
\author{
\IEEEauthorblockN{
Hafija Aktar$^{1}$,
Molly Croes$^{1}$,
Frank S. Menniti$^{2}$,
and Abdeltawab Hendawi$^{1}$
}
\IEEEauthorblockA{
$^{1}$Department of Computer Science\\
$^{2}$Department of Biomedical and Pharmaceutical Sciences,
George \& Anne Ryan Institute for Neuroscience\\
University of Rhode Island, Kingston, RI 02881, USA\\
\{hafijaaktar, molly.croes, mennitifs, hendawi\}@uri.edu
}
}

\maketitle

\begin{abstract}
Accurate assessment of Alzheimer’s disease stages from MRI is important for understanding disease progression and supporting the development of computer-aided analysis tools. Deep learning models have shown promising results for Alzheimer’s disease classification from MRI. However, when evaluated on single datasets, it is difficult to determine whether the observed performance reflects consistent architectural behavior or dataset-specific effects. In this work, we present a multi-dataset benchmarking framework with standardized preprocessing and evaluation. We compare five architectures: ResNet-50, ReXNet-150, Swin Transformer, ViT-Base, and ViT-Large. The models are evaluated across three MRI datasets representing severe class imbalance, train-test distribution mismatch, and large-scale augmentation. The results show broadly consistent performance patterns across datasets, with ViT-Large achieving the best performance on all three datasets and reaching up to 99.84\% accuracy and 99.86\% macro F1. Performance differences between architectures are most pronounced under distribution mismatch and considerably smaller in the augmented-data setting. Across datasets, lower class-level performance is observed primarily among adjacent early-stage categories, particularly Very Mild and Mild impairment. These findings highlight the importance of evaluating deep learning architectures across multiple data conditions rather than relying on single-dataset performance.
\end{abstract}
\begin{IEEEkeywords}
Alzheimer's disease, MRI classification, deep learning,
multi-dataset benchmarking, distribution mismatch, class imbalance,
transformer models, convolutional neural networks
\end{IEEEkeywords}

\section{Introduction}

Alzheimer’s disease (AD) is a progressive neurodegenerative disorder that affects millions of people worldwide and places a growing burden on healthcare systems \cite{wen2020convolutional, basaia2019automated}. Structural MRI provides a non-invasive way to detect disease-related changes in the brain \cite{basaia2019automated}. Accurate staging is important for disease assessment. In MRI-based classification tasks, deep learning models have demonstrated great accuracy, frequently above 95\% on benchmark datasets \cite{wen2020convolutional}. However, these results are often reported on a single dataset. As a result, it remains difficult to determine whether observed performance reflects consistent model behavior or dataset-specific effects. Similar concerns have been observed in external validation studies of artificial intelligence (AI) systems in radiology~\cite{yagis2021effect, yu2022external}. It is difficult to tell whether differences in performance between architectures are consistent across diverse data situations.

MRI datasets have different sizes, class distributions and conventional MRI designs. Synthetic balancing and augmentation can change the properties of the data and cause distributional shifts. However, real-world datasets are often imbalanced \cite{goodfellow2014generative, guan2021domain, lteif2024disease}. Transformer based designs use attention processes for capturing larger dependencies, while convolutional models are based on local spatial inductive biases \cite{50650}. Architectural choices may influence the way models respond to changes in distribution, imbalance, and size of data.

The classification task is made more difficult by the characteristics of Alzheimer's disease. The stages of the disease's progression range from Very Mild to Moderate impairment. Even experts may find it challenging to identify the subtle structural changes that characterize early stages \cite{jack2010hypothetical, frisoni2010clinical}. Consequently, classification difficulty is not uniform across classes, and adjacent early-stage categories may be particularly difficult to distinguish.

Systematic comparisons across multiple datasets are still limited, despite the growing interest in deep learning for classifying Alzheimer's MRI. Thus, it is not clear whether the relative ranking of model performance is consistent across datasets or how much they are sensitive to factors such as train-test distribution mismatch, data augmentation, and class imbalance. This gap leads to the following research question:

\begin{quote}
\textit{Are performance patterns of CNNs and vision transformers for Alzheimer's MRI classification consistent across datasets with different distributional properties, and how are these patterns affected by dataset-specific characteristics?}
\end{quote}

In order to address this question, we present a multi-dataset benchmarking approach that compares transformer-based and convolutional models under standardized preprocessing and evaluation. To facilitate efficient training in each context, the framework incorporates optimization parameters specific to datasets and architectures. We evaluate five benchmark architectures: ViT-Base, ViT-Large, Swin Transformer, ReXNet-150 and ResNet-50. We assess the models across three datasets, large-scale augmentation, synthetic balancing with train-test distribution mismatch and severe class imbalance. This setting enables us to compare the trends in performance across quite different data conditions in a particular evaluation framework.

We use this framework to study how architectural design, model capacity, and dataset characteristics are associated with classification performance. In particular, we examine changes in model ranking, sensitivity to distribution mismatch, and class-level performance across disease stages.

The contributions of this work are:
\begin{itemize}

\item We present a multi-dataset benchmarking methodology with standardized preprocessing and evaluation for systematically comparing deep learning architectures under different data conditions, including class imbalance, train-test distribution mismatch, and large-scale augmentation.

\item We show that model performance patterns are broadly consistent across the evaluated datasets, with ViT-Large achieving the strongest overall performance on all three datasets and Swin Transformer consistently ranking among the top models.

\item We analyze how the magnitude of performance differences changes across dataset conditions and show that architecture-level differences become more pronounced under the distribution-mismatch setting.

\item We identify adjacent early-stage categories, particularly Very Mild and Mild impairment, as the classes with lower and more variable class-level performance across the evaluated models and datasets.

\end{itemize}

The remainder of this paper is organized as follows. Section~\ref{sec:related} reviews prior work on MRI-based Alzheimer's disease classification, transformer models, and dataset shift. Section~\ref{sec:datasets} describes the three MRI datasets and their class distribution characteristics. Section~\ref{sec:methodology} presents the evaluated architectures, fine-tuning strategy, and training configuration. Section~\ref{sec:results} reports the experimental results and cross-dataset analysis. Section~\ref{sec:limitations} discusses the limitations of the study, and Section~\ref{sec:conclusion} concludes the paper.

\section{Related Work}
\label{sec:related}

\begin{table*}[t]
\caption{Representative studies on MRI-based Alzheimer’s disease classification, covering CNN, transformer, and hybrid models, along with multi-dataset benchmarking and dataset shift.}
\label{tab:related_work_comparison}
\centering
\scriptsize
\setlength{\tabcolsep}{4pt}
\renewcommand{\arraystretch}{1.1}
\resizebox{\textwidth}{!}{%
\begin{tabular}{p{2.5cm} p{0.5cm} p{1.9cm} p{2.0cm} L{4.8cm} L{3.6cm}}
\hline
Paper & Year & Data & Model & Key finding & Relevance \\
\hline

He et al.~\cite{he2016deep} 
& 2016 
& ImageNet 
& ResNet 
& Residual learning enables deep, transferable CNNs. 
& CNN baseline \\

Han et al.~\cite{han2021rethinking} 
& 2021 
& ImageNet 
& ReXNet 
& Efficient CNN design reduces representational bottlenecks. 
& Efficient CNN \\

Dosovitskiy et al.~\cite{50650} 
& 2021 
& ImageNet 
& ViT 
& Patch-based transformers achieve strong performance at scale. 
& ViT basis \\

Liu et al.~\cite{liu2021swin} 
& 2021 
& ImageNet 
& Swin 
& Hierarchical shifted-window attention improves efficiency and representation. 
& Swin baseline \\

Basaia et al.~\cite{basaia2019automated} 
& 2019 
& ADNI, clinical 
& 3D CNN 
& Single structural MRI scans can support AD classification. 
& CNN AD study \\

Wen et al.~\cite{wen2020convolutional} 
& 2020 
& ADNI, AIBL, OASIS 
& CNN, SVM 
& Highlights reproducibility issues and limited cross-dataset generalization. 
& Benchmarking gap \\

Bron et al.~\cite{bron2021cross} 
& 2021 
& ADNI, clinical 
& CNN, SVM 
& Performance decreases when models are evaluated across independent cohorts. 
& Cross-cohort generalization \\

Jang \& Hwang~\cite{jang2022m3t} 
& 2022 
& ADNI, AIBL, OASIS 
& M3T 
& Hybrid modeling captures multi-plane MRI information. 
& Transformer AD model \\

Xin et al.~\cite{xin2023cnn} 
& 2023 
& ADNI, AIBL 
& CNN + Swin 
& Hybrid CNN--Swin model is evaluated with external validation. 
& Hybrid trend \\

Kushol et al.~\cite{kushol2023effects} 
& 2023 
& Multi-site MRI 
& DL models 
& Scanner variation degrades classification performance. 
& Dataset shift \\

Zech et al.~\cite{zech2018variable} 
& 2018 
& Multi-site medical imaging 
& DL models 
& Generalization performance varies substantially across institutions. 
& Dataset shift \\

Castro et al.~\cite{castro2020causality} 
& 2020 
& Medical imaging 
& DL models 
& Highlights the influence of dataset bias and causal factors on generalization. 
& Dataset bias \\

Chen et al.~\cite{chen2024longformer} 
& 2024 
& ADNI, OASIS, AIBL 
& LongFormer 
& Longitudinal modeling improves classification performance. 
& Temporal modeling \\

Alp et al.~\cite{alp2024joint} 
& 2024 
& ADNI 
& Joint Transformer 
& Attention-based 3D MRI modeling supports AD classification. 
& Transformer AD model \\

Song et al.~\cite{song2022reliability} 
& 2022 
& Multi-cohort MRI 
& DL model 
& Multi-cohort, multi-vendor, and multi-protocol validation reveals generalization limitations. 
& External validation \\

Yagis et al.~\cite{yagis2021effect} 
& 2021 
& Brain MRI 
& 2D CNN 
& Data leakage can inflate MRI classification performance. 
& Leakage risk \\

\hline
This study 
& 2026 
& 3 MRI datasets 
& CNN, ViT, Swin 
& Compares architecture performance patterns across datasets with different distributional characteristics. 
& Multi-dataset benchmarking \\
\hline

\end{tabular}%
}
\end{table*}

Deep learning has become a widely used approach for classifying Alzheimer’s disease (AD) from structural MRI, with advances in both convolutional and transformer-based models. Early work demonstrated that 3D CNNs can classify AD from a single MRI scan under controlled conditions \cite{basaia2019automated}. Later studies highlighted an important limitation of this literature: results are often difficult to compare because of differences in preprocessing, data splits, and evaluation protocols~\cite{wen2020convolutional}. Recent reviews also show increasing interest in comparing CNN and transformer-based approaches for Alzheimer’s disease classification from MRI \cite{bravo2024systematic}. Standard CNN backbones such as ResNet and ReXNet remain useful baselines because of their spatial inductive bias and computational efficiency \cite{he2016deep, han2021rethinking}.

Transformer-based models represent a different modeling paradigm, employing attention to capture long-range dependencies. Vision Transformers (ViT) performed well at scale, and Swin Transformer introduced hierarchical attention with better computational efficiency \cite{50650, liu2021swin}. Training strategies for data-efficient transformers have been explored to improve optimization when large-scale labeled data are limited \cite{touvron2021training}. CNN-transformer and transformer-based architectures have been explored in the AD domain to capture local anatomical structure and broader spatial relationships \cite{xin2023cnn, jang2022m3t}. Recently, transformer-based attention mechanisms have also been used for 3D brain MRI classification \cite{alp2024joint}. Although these approaches are promising, many studies evaluate a limited number of architectures on individual datasets which makes it difficult to determine whether reported improvements are due to the benefits of the architecture or the effects of the dataset.

These concerns become particularly important when model behavior is compared across datasets or evaluated on external cohorts. Previous studies have shown that strong internal performance may not translate to independent populations. Bron et al.~\cite{bron2021cross} reported reduced performance under cross-cohort evaluation, while Song et al.~\cite{song2022reliability} emphasized the importance of multi-cohort, multi-vendor, and multi-protocol validation. More broadly, medical imaging studies associate generalization failures with dataset shift caused by differences in acquisition protocols, population characteristics, and hidden biases \cite{zech2018variable, castro2020causality, guan2021domain}. In MRI, scanner-specific characteristics can further affect model performance across sites \cite{glocker2019machine, kushol2023effects, lteif2024disease}.

Data construction and augmentation can also influence model evaluation. Generative adversarial networks provide a general framework for generating synthetic data \cite{goodfellow2014generative}, and synthetic samples are often used to expand training data or reduce class imbalance. However, synthetic balancing and heavy augmentation can alter the training distribution and may not fully represent real clinical conditions. Data leakage is another important concern in MRI classification, particularly when related or visually similar samples occur across training and testing splits, which can inflate reported performance~\cite{yagis2021effect}. In addition, differences in preprocessing, imbalance handling, optimization, and training strategies can complicate performance comparisons across studies \cite{wen2020convolutional, 50650, touvron2021training}. For a clearer comparison among existing approaches, Table~\ref{tab:related_work_comparison} summarizes representative studies in terms of datasets, models, and key findings.

Although prior studies have evaluated CNN, transformer, and hybrid models for Alzheimer’s MRI classification, differences in datasets, preprocessing, training settings, and evaluation protocols limit direct comparison across studies. Existing work has also focused largely on reporting performance within individual datasets or on external validation of a particular model, leaving comparatively less evidence on how the relative behavior of multiple architectures changes across different dataset conditions. We therefore evaluate representative CNN and transformer architectures across three datasets using consistent preprocessing and evaluation, while adapting training strategies to the characteristics of each dataset. This design allows us to examine whether architecture rankings and class-level performance patterns remain broadly consistent under severe class imbalance, train-test distribution mismatch, and large-scale augmentation.

\section{Datasets}
\label{sec:datasets}

\begin{figure*}[!t]
\centering
\includegraphics[width=\textwidth]{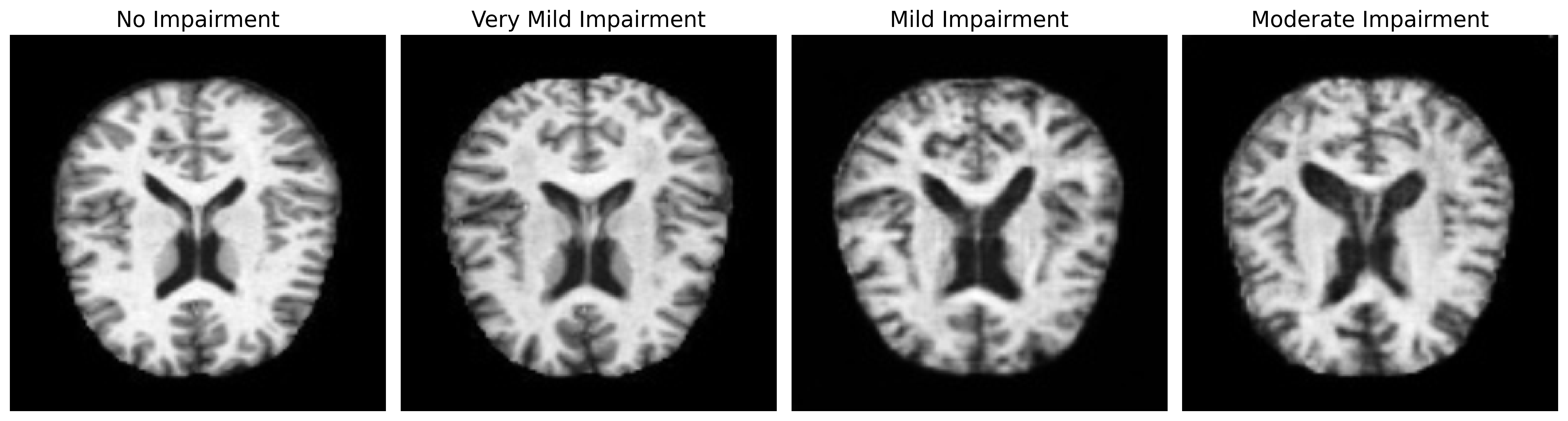}
\caption{Representative axial MRI slices illustrating the four diagnostic categories used in this study.}
\label{fig:dataset2_examples}
\end{figure*}

We conduct experiments on three publicly available structural MRI datasets for AD stage classification. Each dataset is composed of brain MRI images annotated with four diagnostic stages corresponding to increasing levels of cognitive impairment. The categories for Datasets~I and~III are Non
Demented, Very Mild Demented, Mild Demented, and Moderate Demented, while Dataset~II uses No Impairment, Very Mild Impairment, Mild Impairment, and Moderate Impairment. Despite differences in naming, these labels represent the same progression from cognitively normal to advanced stages.
Representative MRI slices for the four diagnostic categories are shown in Fig.~\ref{fig:dataset2_examples}.

The three datasets provide complementary evaluation settings with different construction methods, scales, and class distributions. Using multiple datasets allows us to examine whether the relative behavior of CNNs and vision transformers remains stable when the properties of the data change, rather than drawing conclusions from a single benchmark. Dataset~I consists of real MRI images with severe class imbalance,
allowing us to examine model behavior when some disease stages are represented by very few samples. Dataset~II contains a balanced training set composed of real and synthetically generated images using Wasserstein GAN with Gradient Penalty (WGAN-GP), together with a highly imbalanced
test set, creating a challenging train--test distribution mismatch. Dataset~III consists of augmented MRI images generated through systematic transformations and provides a substantially larger augmented-data setting. This setting allows us to examine whether architecture-level performance differences remain visible when more augmented samples are
available. Together, these datasets enable us to study model behavior under severe class imbalance, distribution mismatch, and large-scale augmentation. Table~\ref{tab:dataset_characteristics} summarizes their main characteristics and evaluation focus.

For Datasets~I and~III, the originally provided training and test images are combined and repartitioned using stratified splits of approximately 70\%, 15\%, and 15\% for training, validation, and testing, respectively. For Dataset~II, the original test set is preserved to maintain the intended imbalanced evaluation setting, while the provided training set is divided into training and validation subsets using an 80\%--20\% stratified split. Stratified sampling is used for all newly created partitions to preserve the corresponding class distributions as closely as possible.

\begin{table*}[!t]
\caption{Key characteristics and evaluation focus of the three MRI datasets.}
\label{tab:dataset_characteristics}
\centering
\small
\setlength{\tabcolsep}{5pt}
\renewcommand{\arraystretch}{1.15}
\resizebox{\textwidth}{!}{%
\begin{tabular}{p{1.6cm} p{2.2cm} p{2.8cm} p{3.4cm} p{4.6cm}}
\hline
Dataset & Total Images & Data Type & Key Property & Evaluation Focus \\
\hline
Dataset I 
& 6{,}400 
& Real MRI images 
& Severe class imbalance 
& Impact of severe class imbalance on model performance \\

Dataset II 
& 11{,}519 
& Real and synthetic MRI images 
& Balanced train, imbalanced test 
& Robustness under train-test distribution mismatch \\

Dataset III 
& 33{,}984 
& Augmented MRI images 
& Large-scale augmentation with moderate imbalance
& Effect of large-scale augmentation on classification performance \\
\hline
\end{tabular}%
}
\end{table*}

\subsection{Dataset Descriptions}

We now include a detailed description of each dataset, emphasizing its size, composition, and class distribution. These datasets are chosen to represent diverse data conditions, including real-world class imbalance, synthetic
balancing with distribution mismatch, and large-scale augmentation. This combination provides complementary test conditions for comparing the five architectures across substantially different dataset characteristics.

\begin{table*}[!t]
\caption{Per-class distribution across all three datasets after stratified splitting.}
\label{tab:dataset_combined}
\centering
\small
\setlength{\tabcolsep}{8pt}
\renewcommand{\arraystretch}{1.2}
\resizebox{\textwidth}{!}{%
\begin{tabular}{lccccccccc}
\toprule
& \multicolumn{3}{c}{Dataset I}
& \multicolumn{3}{c}{Dataset II}
& \multicolumn{3}{c}{Dataset III} \\
\cmidrule(lr){2-4}
\cmidrule(lr){5-7}
\cmidrule(lr){8-10}

Class 
& Train & Val & Test 
& Train & Val & Test 
& Train & Val & Test \\
\midrule
Non / No Impairment
& 2{,}286 & 457 & 457 
& 2{,}042 & 518 & 640 
& 6{,}720 & 1{,}440 & 1{,}440 \\

Very Mild 
& 1{,}568 & 320 & 320 
& 2{,}041 & 519 & 448 
& 6{,}272 & 1{,}344 & 1{,}344 \\

Mild 
& 672 & 128 & 128 
& 2{,}048 & 512 & 179 
& 6{,}272 & 1{,}344 & 1{,}344 \\

Moderate 
& 46 & 9 & 9 
& 2{,}061 & 499 & 12 
& 4{,}524 & 970 & 970 \\
\midrule
\textbf{Split Total}
& \textbf{4{,}572} & \textbf{914} & \textbf{914}
& \textbf{8{,}192} & \textbf{2{,}048} & \textbf{1{,}279}
& \textbf{23{,}788} & \textbf{5{,}098} & \textbf{5{,}098} \\
\bottomrule
\end{tabular}%
}
\end{table*}
\subsubsection{Dataset I: Alzheimer MRI Disease Classification Dataset}
This dataset, available on Hugging Face~\cite{alzheimer_mri_dataset}, contains brain MRI images labelled by four diagnostic categories: Non Demented, Very Mild Demented, Mild Demented, and Moderate Demented. The original dataset consists of 5{,}120 training images and 1{,}280 test images, for a total of 6{,}400 samples. After combining the original splits and applying stratified partitioning, the dataset comprises 4{,}572 training, 914 validation and 914 test images. The dataset has a highly imbalanced class distribution, with Non Demented images forming the majority and only 64 Moderate Demented samples in total (46 train, 9 validation, and 9 test). This extreme imbalance provides a challenging class-imbalanced setting for model evaluation.

\subsubsection{Dataset II: Best Alzheimer MRI Dataset}
This dataset, obtained from Kaggle~\cite{chugh2023best_alzheimer_mri}, contains axial brain MRI images across four diagnostic stages: No Impairment, Very Mild Impairment, Mild Impairment, and Moderate Impairment. To overcome the data scarcity and class imbalance issues, the dataset was generated by combining real MRI scans and synthetically generated images using Wasserstein GAN with Gradient Penalty (WGAN-GP). The training set consists of 10,240 images, split approximately equally among the classes, with 2,560 images per class prior to partitioning. After stratified partitioning there are 8{,}192 training images and 2{,}048 validation images. The original test set of 1{,}279 images is preserved. In contrast to the balanced training data, the test set is highly imbalanced, with 640 No Impairment samples and only 12 Moderate Impairment samples. This train-test asymmetry creates a difficult evaluation setting under distribution mismatch. In this study, this mismatch refers primarily to the difference in class proportions between the balanced training set and the highly imbalanced test set, together with differences introduced by the synthetic composition of the training data.

\subsubsection{Dataset III: Augmented Alzheimer MRI Dataset}
This Kaggle dataset~\cite{uraninjo2023augmented_alzheimer_mri} contains Alzheimer’s MRI images in four classes: Non Demented, Very Mild Demented, Mild Demented, and Moderate Demented. It includes original and augmented images generated from an existing Alzheimer's MRI dataset. In this study, only the augmented images are used. Because these samples are generated from a small number of original images, they may introduce redundancy across dataset splits and inflate performance if highly similar samples appear in both training and testing partitions~\cite{tampu2022inflation, young2025data}. Thus, Dataset~III results should be interpreted as augmented-data performance rather than evidence of clinical generalization. After stratified partitioning, the dataset contains 23{,}788 training, 5{,}098 validation, and 5{,}098 test images, totaling 33{,}984 samples. Dataset III is much larger than Dataset I and has moderate class imbalance, with similar class proportions across splits.

\subsection{Class Distribution}

The three datasets present different imbalance patterns that can influence both model training and evaluation. These differences are important in multi-class Alzheimer’s disease classification because disease stages are not equally represented, which can affect how reliably models learn and distinguish stage-specific features. The per-class distributions across all datasets are summarized in Table~\ref{tab:dataset_combined}.

Dataset I has a very imbalanced setting, with only 46 training samples for the Moderate Demented class compared to 2{,}286 Non Demented samples. This extreme imbalance introduces a strong bias to the majority classes in the training phase, making it especially challenging to accurately identify minority classes. This makes models trained on this dataset susceptible to overfitting of dominant patterns, and may hinder their generalization to underrepresented disease stages.

Dataset II presents a different challenge. While the training data is fairly balanced across all classes, the test set is very imbalanced with 12 samples of Moderate Impairment and 640 samples of No Impairment. This train-test distribution mismatch provides a challenging evaluation setting in which models trained on balanced data are evaluated under a highly imbalanced class distribution.

Dataset III represents a moderately imbalanced setting across all splits. Although class sizes are more evenly distributed than in Dataset I, the Moderate Demented class is still the smallest. The class distribution in train, validation and test sets is similar, which provides a consistent environment for training.

Together, the datasets represent three complementary evaluation settings: severe class imbalance in Dataset~I, train--test distribution mismatch in Dataset~II, and large-scale augmentation with more moderate class imbalance in Dataset~III. These settings allow us to examine whether architecture performance remains consistent as the underlying data conditions change.

\begin{figure*}[t]
\centering
\includegraphics[width=0.7\textwidth,height=0.8\textheight,keepaspectratio]{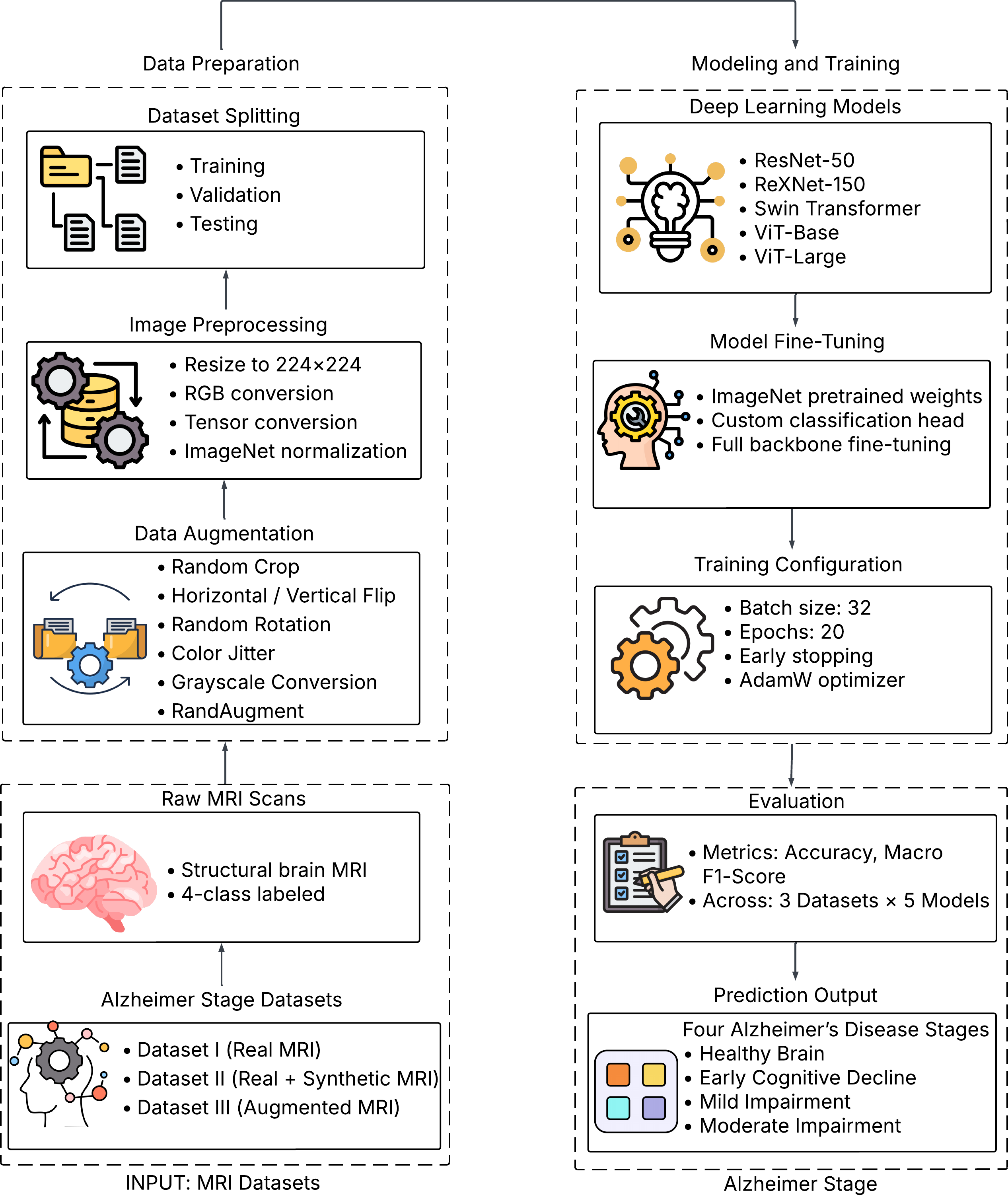}
\caption{Overview of the experimental pipeline for Alzheimer's disease MRI stage classification. MRI datasets are preprocessed and augmented before being used to train multiple deep learning models. The models are fine-tuned and evaluated using classification metrics to predict Alzheimer's disease stages.}
\label{fig:pipeline}
\end{figure*}


\subsection{MRI Data Preparation}

All MRI images are processed with a standardized pipeline to ensure consistent input representation across datasets and architectures. The images are scaled to $224 \times 224$ pixels to match the requirements of the pretrained model, converted to PyTorch tensors, and scaled from $[0, 255]$ to $[0, 1]$. The images are normalized with ImageNet channel-wise mean and standard deviation to align with the pretrained feature distributions and allow efficient transfer learning. MRI images are grayscale, so each image is replicated in three channels to match the pretrained models' input format. This preprocessing pipeline reduces input-related variability when comparing architectural and training configurations.

Data augmentation is only performed on the training set to improve generalization and reduce overfitting. The augmentation pipeline incorporates random cropping, horizontal and vertical flipping, random rotations, color jitter, grayscale conversion, and RandAugment to introduce spatial and intensity-based variability. Although MRI images are grayscale, color jitter and grayscale conversion are retained in the implemented pipeline after channel replication, where they function as intensity-level perturbations rather than true color transformations.  Validation and test images are processed deterministically only to ensure consistent and reproducible evaluation. The complete pipeline for preprocessing, augmentation, model training and evaluation is represented in Figure~\ref{fig:pipeline}. The figure illustrates how the three MRI datasets are first standardized through resizing, channel conversion, normalization, and training-only augmentation, followed by model fine-tuning and evaluation across the five benchmark architectures.

\section{Methodology}
\label{sec:methodology}

The accurate staging of Alzheimer’s disease from structural MRI requires models that can capture subtle and progressive anatomical changes across different stages of cognitive impairment. Convolutional neural networks (CNNs) have shown their capacity to learn hierarchical spatial representations of brain structures~\cite{sarraf2016deepad,wen2020convolutional,basaia2019automated,bae2020identification} and transformer-based models can capture long-range dependencies using attention mechanisms~\cite{50650,liu2021swin,chen2024transunet,hatamizadeh2022unetr}.

Despite these advancements, systematic comparisons among CNN and transformer-based models are still limited, especially on datasets with different distribution characteristics.
To address this, we evaluate representative architectures from both families under consistent preprocessing and evaluation settings, while adapting training strategies to each dataset. This design supports comparison of architectural performance by standardizing preprocessing and evaluation while allowing dataset-specific training adaptations.
\begin{table*}[t]
\caption{Architectural characteristics of the evaluated deep learning models.}
\label{tab:model_summary}
\centering
\small
\setlength{\tabcolsep}{6pt}
\renewcommand{\arraystretch}{1.1}
\begin{tabular}{lccccc}
\toprule
Model & Architecture Type & Depth & Patch Size & Attention Heads & Parameters \\
\midrule
ResNet-50       & CNN                      & 50  & --            & --               & 25M  \\
ReXNet-150       & CNN                      & 150 & --            & --               & 16M  \\
Swin Transformer & Hierarchical Transformer & 24  & $4\times4$    & 3 / 6 / 12 / 24  & 88M  \\
ViT-Base         & Vision Transformer       & 12  & $16\times16$  & 12               & 86M  \\
ViT-Large        & Vision Transformer       & 24  & $16\times16$  & 16               & 307M \\
\bottomrule
\end{tabular}
\end{table*}

\subsection{Benchmark Architectures}

We evaluate five representative architectures: ResNet-50, ReXNet-150, Swin Transformer, ViT-Base, and ViT-Large. These models cover both convolutional and transformer-based approaches, enabling comparison between local spatial feature extraction and global attention-based modeling for Alzheimer's disease MRI classification. They also differ in depth and parameter size, allowing us to examine performance patterns associated with model capacity. This selection provides a balanced benchmark of standard CNNs, efficient CNNs, hierarchical transformers, and Vision Transformer variants. Their key architectural characteristics are summarized in Table~\ref{tab:model_summary}.

\subsubsection{ResNet-50}

ResNet-50~\cite{he2016deep} is a deep convolutional neural network that uses residual shortcut connections to improve gradient flow and reduce the vanishing-gradient problem in very deep models. Instead of learning a direct mapping, each residual block learns a residual function, which makes optimization more stable and allows the network to benefit from increased depth. The architecture consists of an initial convolutional stem followed by four stages of bottleneck residual blocks, where each block uses $1\times1$, $3\times3$, and $1\times1$ convolutions to reduce, process, and restore feature dimensions efficiently. ResNet-50 has also been applied directly to Alzheimer’s MRI classification, supporting its use as a relevant CNN baseline for this task~\cite{fulton2019classification}. In this study, ResNet-50 is used as the main CNN reference model because its residual design provides a widely adopted and robust baseline for comparing convolutional feature learning with transformer-based architectures in MRI classification.

\subsubsection{ReXNet-150}

ReXNet-150~\cite{han2021rethinking} is an efficient convolutional architecture that aims at improving the representational capacity by reducing channel bottlenecks in CNN feature maps. The model progressively expands the channels’ size along the layers, so the feature representations are not excessively compressed in the forward propagation. This design is motivated by the observation that narrow intermediate representations may limit the information flow and reduce the expressive power of lightweight CNNs. ReXNet is also designed with efficient-CNN design principles similar to MobileNet-style architectures, but with a special focus on reducing the representational bottlenecks through progressive channel expansion~\cite{howard2017mobilenets, sandler2018mobilenetv2}. In this work, we use ReXNet-150 as an efficient CNN baseline to evaluate if better channel allocation and parameter-efficient convolutional design can bring benefits for MRI-based Alzheimer’s disease classification.

\subsubsection{Swin Transformer}

The Swin Transformer~\cite{liu2021swin} is a hierarchical vision transformer designed to improve the efficiency of standard self-attention. Instead of computing attention globally over all image patches, it restricts self-attention to non-overlapping local windows, which substantially reduces computational complexity. To allow information exchange across neighboring windows, Swin introduces a shifted-window mechanism in which the window partition is shifted between successive layers. This enables cross-window interactions while preserving efficient local attention. The architecture also progressively reduces spatial resolution and increases feature dimensionality, producing a multi-scale representation similar to convolutional feature hierarchies. Such hierarchical and window-based modeling has made Swin-based architectures useful in medical image analysis, including 3D medical representation learning and MRI-based segmentation tasks~\cite{tang2022self, hatamizadeh2021swin}. In Alzheimer’s disease analysis, Swin-based hybrid models have also been explored for structural MRI diagnosis~\cite{xin2023cnn}. In this study, the Swin Transformer is included to evaluate whether hierarchical attention can capture both local anatomical variations and broader structural context in MRI-based Alzheimer’s disease classification.

\subsubsection{Vision Transformer (ViT-Base)}

The Vision Transformer (ViT)~\cite{50650} encodes an image as a sequence of fixed-size patches, instead of convolutional feature maps. The input image in ViT-Base is split into $16\times16$ non-overlapping patches, which are then flattened, linearly projected to embedding vectors and the positional encodings are added to them. The resulting patch sequence is then processed by stacked transformer encoder blocks consisting of multi-head self-attention and feed-forward layers. This design enables each patch token to attend to other parts of the image during representation learning. This global attention at the patch level is beneficial for stage classification as the structural difference between the Alzheimer stages can be present in many brain regions instead of one localized region. Recently, ViT-based feature extraction has also been used in Alzheimer’s MRI studies for 3D MRI classification~\cite{alp2024joint}. The ViT-Base has 12 encoder layers, 12 attention heads and ~86M parameters. In this study, it serves as the baseline transformer model for assessing global self-attention in MRI-based Alzheimer’s disease classification.
\begin{table*}[t]
\caption{Model-specific optimization settings across the three datasets. A single learning rate indicates that the same rate is applied to the backbone and classification head.}
\label{tab:training_config}
\centering
\footnotesize
\setlength{\tabcolsep}{4pt}
\begin{tabular}{lcccccc}
\toprule
& \multicolumn{2}{c}{Dataset~I}
& \multicolumn{2}{c}{Dataset~II}
& \multicolumn{2}{c}{Dataset~III} \\
\cmidrule(lr){2-3}
\cmidrule(lr){4-5}
\cmidrule(lr){6-7}

Model
& \makecell{Learning Rate\\(Backbone / Head)}
& \makecell{Weight\\Decay}
& \makecell{Learning\\Rate}
& \makecell{Weight\\Decay}
& \makecell{Learning Rate\\(Backbone / Head)}
& \makecell{Weight\\Decay} \\
\midrule

ResNet-50
& $3{\times}10^{-4}$ / $1{\times}10^{-3}$
& $10^{-4}$
& $3{\times}10^{-4}$
& $10^{-4}$
& $1{\times}10^{-4}$
& $10^{-4}$ \\

ReXNet-150
& $3{\times}10^{-4}$ / $1{\times}10^{-3}$
& $10^{-4}$
& $1{\times}10^{-4}$
& $10^{-4}$
& $1{\times}10^{-4}$
& $10^{-4}$ \\

Swin Transformer
& $5{\times}10^{-5}$ / $1{\times}10^{-4}$
& $0.05$
& $5{\times}10^{-5}$
& $10^{-4}$
& $5{\times}10^{-5}$
& $10^{-4}$ \\

ViT-Base
& $5{\times}10^{-5}$ / $1{\times}10^{-4}$
& $0.05$
& $1{\times}10^{-4}$
& $10^{-4}$
& $5{\times}10^{-6}$ / $5{\times}10^{-5}$
& $10^{-4}$ \\

ViT-Large
& $3{\times}10^{-5}$ / $1{\times}10^{-4}$
& $0.05$
& $1{\times}10^{-4}$
& $10^{-4}$
& $2{\times}10^{-6}$ / $5{\times}10^{-5}$
& $10^{-4}$ \\
\bottomrule
\end{tabular}
\end{table*}

\subsubsection{Vision Transformer (ViT-Large)}

ViT-Large~\cite{50650} follows the same patch-based transformer design as ViT-Base but increases model capacity substantially. It uses 24 encoder layers, 16 attention heads, and approximately 307M parameters, compared with 12 encoder layers, 12 attention heads, and approximately 86M parameters in ViT-Base. Larger vision transformers can learn more expressive patch-level interactions, and prior work on scaling vision transformers shows that increasing model size, data, and compute can improve visual representation quality~\cite{zhai2022scaling}. In the context of Alzheimer’s MRI classification, this larger capacity may help capture subtle stage-related structural differences distributed across brain regions. In this study, comparing ViT-Large with ViT-Base allows us to assess whether increased transformer capacity improves classification performance under different dataset conditions.

\subsection{Model Fine-Tuning Strategy}
All models are initialized with ImageNet-pretrained weights and fine-tuned for four-class Alzheimer’s disease MRI stage classification~\cite{tajbakhsh2016convolutional, raghu2019transfusion}. The original classification head for each architecture is replaced or adapted to predict the four diagnostic categories.

For ResNet-50 and ReXNet-150, we replace the final fully connected layer with a linear classifier that maps the learned feature representation to the four output classes. For transformer-based architectures, the classification head is customized according to dataset characteristics. We keep the default pretrained linear classifier on Dataset~I. For Datasets~II and~III, a more expressive custom head is used to support optimization under class imbalance and larger training sets. For ViT-Base and ViT-Large, this head consists of LayerNorm followed by two fully connected layers with dropout and non-linear activations. The Swin Transformer head consists of Layer Normalization, a projection layer, dropout, and a final output layer.

Training strategies are adapted to dataset size and class distribution. On Dataset~I, we perform a two-epoch warm-up phase where only the classification head is trained and the pretrained backbone is frozen. This is followed by end-to-end fine-tuning of all model parameters. On Datasets~II and~III, all parameters are fine-tuned from the start, allowing pretrained representations to adapt directly to MRI-specific features.

For transformer-based models, we use a split learning-rate strategy on Datasets~I and~III: the pretrained backbone uses a lower learning rate and the classification head uses a higher learning rate. This preserves pretrained representations while allowing newly initialized task-specific layers to adapt faster. For Dataset~II, we use a single learning rate for all parameters.

\subsection{Loss Function and Optimization}

All models are trained using cross-entropy loss with label smoothing to improve calibration and reduce overconfident predictions~\cite{szegedy2016rethinking}. A smoothing value of 0.02 is used for Dataset~I, while 0.1 is used for Datasets~II and~III to provide stronger regularization for the larger training sets.

Class imbalance is handled with dataset-specific strategies. For Dataset~I, which has severe imbalance, we use a tempered \textit{WeightedRandomSampler}, with sampling weights computed as inverse class frequency raised to $\alpha=0.5$. No class weights are added to the loss in this setting. For Dataset~II, inverse-frequency class weights are incorporated directly into the loss function. Dataset~III has only moderate imbalance and is trained without explicit imbalance correction.

Models are optimized using AdamW across all architectures and
datasets~\cite{kingma2014adam}. AdamW extends Adam by decoupling weight decay from the adaptive gradient update, providing explicit regularization during fine-tuning~\cite{loshchilov2019iclr-decoupled}. Using the same optimizer across all configurations also reduces optimizer-related variation when comparing the evaluated architectures. The complete learning-rate and weight-decay settings are summarized in Table~\ref{tab:training_config}.

Learning-rate scheduling is adapted by dataset. Cosine annealing is used for Dataset~I~\cite{loshchilov2016sgdr}, while \textit{ReduceLROnPlateau} is used for Datasets~II and~III, reducing the learning rate when validation macro F1-score plateaus. To improve stability and efficiency, mixed precision training, gradient accumulation, and gradient clipping with a maximum norm of 1.0 are applied on Datasets~I and~III.

\subsection{Training Configuration}

We train all models for a maximum of 20 epochs with a mini-batch size of 32. Validation performance is monitored during training, and early stopping is used to mitigate overfitting. Training stops if the validation score does not improve for 4 epochs on Dataset~I or 5 epochs on Datasets~II and~III. The validation metric is selected based on dataset characteristics: Balanced Accuracy is used for Dataset~I because of severe class imbalance, while Macro F1 is used for Datasets~II and~III. For each experiment, the model with the best validation score is selected for final testing. Model-specific optimization settings are listed in Table~\ref{tab:training_config}. Table~\ref{tab:global_training} summarizes the hyperparameters shared across models. Gradient accumulation is used on Datasets~I and~III as described in Table~\ref{tab:dataset_training_config}. Dataset-specific training settings are also listed in Table~\ref{tab:dataset_training_config}. These tables summarize the input resolution, batch size, epoch limit, learning-rate settings, imbalance-handling strategies, and early-stopping criteria used across experiments.

\begin{table}[!ht]
\caption{Common training hyperparameters used across the evaluated models.}
\label{tab:global_training}
\centering
\small
\begin{tabular}{lc}
\toprule
Hyperparameter       & Value \\
\midrule
Input Resolution     & $224\times224$ \\
Mini-Batch Size      & 32 \\
Max Epochs           & 20 \\
Loss Function        & Cross-Entropy (Label Smoothing) \\
Gradient Clipping    & 1.0 (Datasets~I \& III) \\
\bottomrule
\end{tabular}
\end{table}

\begin{table*}[t]
\caption{Dataset-specific training configurations. ReduceLROnPlateau reduces the learning rate
by a factor of 0.5. WeightedRandomSampler applies tempered inverse-frequency weights
with $\alpha = 0.5$.}
\label{tab:dataset_training_config}
\centering
\small
\begin{tabular}{lccc}
\toprule
Setting                  & Dataset~I                                     & Dataset~II            & Dataset~III           \\
\midrule
LR Scheduler             & Cosine Annealing                              & ReduceLROnPlateau     & ReduceLROnPlateau     \\
Early Stop Monitor       & Balanced Accuracy                             & Macro F1              & Macro F1              \\
Early Stop Patience      & 4                                             & 5                     & 5                     \\
Label Smoothing          & 0.02                                          & 0.10                  & 0.10                  \\
Imbalance Strategy       & WeightedRandomSampler ($\alpha{=}0.5$)        & Class-weighted loss   & None                  \\
Gradient Accumulation    & 2 steps                                       & None                  & 4 steps               \\
Effective Batch Size     & 64                                            & 32                    & 128                   \\
Head Warmup              & 2 epochs                                      & None                  & None                  \\
\bottomrule
\end{tabular}
\end{table*}

\section{Experimental Results}
\label{sec:results}

We evaluate the proposed benchmarking framework by comparing model performance across the three MRI datasets under the standardized preprocessing and evaluation setup described in the previous section. Models are evaluated using accuracy and macro-averaged F1-score. Accuracy provides an overall measure of correct classification, while macro F1 gives equal importance to each class and is therefore useful for assessing performance when class distributions are imbalanced.

\subsection{Model Performance Across Datasets}

Table~\ref{tab:overall_results} summarizes the overall performance of the five benchmark models across the three datasets using accuracy and macro F1-score. The datasets represent severe class imbalance in Dataset~I, train-test distribution mismatch in Dataset~II, and augmented data with moderate imbalance in Dataset~III. To provide a more detailed view of class-level behavior, Table~\ref{tab:perclass_f1} reports the per-class F1-scores for all models across the three datasets.

Across datasets, ViT-Large and Swin Transformer achieve the strongest performance. Performance patterns are broadly consistent, although dataset-specific factors affect the size of performance gaps, especially on Dataset~II where sensitivity to distribution mismatch varies substantially across models.

\begin{table*}[!t]
\caption{Performance comparison of the five benchmark architectures across the three datasets. Macro F1 is reported to account for class imbalance.}
\label{tab:overall_results}
\centering
\small
\begin{tabular}{lcccccc}
\toprule
 & \multicolumn{2}{c}{Dataset~I} & \multicolumn{2}{c}{Dataset~II} & \multicolumn{2}{c}{Dataset~III} \\
\cmidrule(lr){2-3}\cmidrule(lr){4-5}\cmidrule(lr){6-7}
Model & Acc (\%) & Macro F1 (\%) & Acc (\%) & Macro F1 (\%) & Acc (\%) & Macro F1 (\%) \\
\midrule
ViT-Large         & \textbf{98.47} & \textbf{98.85} & \textbf{97.81} & \textbf{98.30} & \textbf{99.84} & \textbf{99.86} \\
Swin Transformer  & 98.36 & 98.75 & 97.19 & 97.60 & 99.57 & 99.60 \\
ReXNet-150        & 97.81 & 98.38 & 95.70 & 95.80 & 99.41 & 99.46 \\
ResNet-50         & 97.48 & 97.79 & 94.06 & 93.00 & 98.53 & 98.64 \\
ViT-Base          & 97.16 & 98.15 & 91.30 & 86.00 & 98.35 & 98.50 \\
\bottomrule
\end{tabular}
\end{table*}

\begin{table*}[!t]
\caption{Per-class F1-scores of the evaluated models across the three datasets.}
\label{tab:perclass_f1}
\centering
\scriptsize
\setlength{\tabcolsep}{3.2pt}
\renewcommand{\arraystretch}{1.1}

\resizebox{\textwidth}{!}{%
\begin{tabular}{lcccccccccccc}
\toprule
& \multicolumn{4}{c}{Dataset I}
& \multicolumn{4}{c}{Dataset II}
& \multicolumn{4}{c}{Dataset III} \\
\cmidrule(lr){2-5}
\cmidrule(lr){6-9}
\cmidrule(lr){10-13}

Model
& \makecell{Non\\Demented}
& \makecell{Very Mild\\Demented}
& \makecell{Mild\\Demented}
& \makecell{Moderate\\Demented}
& \makecell{No\\Impairment}
& \makecell{Very Mild\\Impairment}
& \makecell{Mild\\Impairment}
& \makecell{Moderate\\Impairment}
& \makecell{Non\\Demented}
& \makecell{Very Mild\\Demented}
& \makecell{Mild\\Demented}
& \makecell{Moderate\\Demented} \\
\midrule

ViT-Large
& 0.9881 & 0.9779 & 0.9881 & 1.0000
& 0.9820 & 0.9710 & 0.9800 & 1.0000
& 0.9976 & 0.9978 & 0.9989 & 1.0000 \\

Swin Transformer
& 0.9857 & 0.9798 & 0.9846 & 1.0000
& 0.9790 & 0.9660 & 0.9600 & 1.0000
& 0.9934 & 0.9914 & 0.9989 & 1.0000 \\

ReXNet-150
& 0.9827 & 0.9684 & 0.9841 & 1.0000
& 0.9600 & 0.9470 & 0.9690 & 0.9570
& 0.9903 & 0.9899 & 0.9981 & 1.0000 \\

ResNet-50
& 0.9837 & 0.9666 & 0.9612 & 1.0000
& 0.9600 & 0.9200 & 0.9200 & 0.9200
& 0.9799 & 0.9758 & 0.9903 & 0.9995 \\

ViT-Base
& 0.9723 & 0.9616 & 0.9922 & 1.0000
& 0.9300 & 0.8900 & 0.9100 & 0.7200
& 0.9730 & 0.9700 & 0.9960 & 1.0000 \\

\bottomrule
\end{tabular}%
}
\end{table*}

\subsection{Results on Dataset~I}

Dataset~I shows severe class imbalance across diagnostic categories. The test set consists of 914 samples, where \textit{Non Demented} (457 samples) and \textit{Very Mild Demented} (320 samples) are the majority classes, while \textit{Moderate Demented} has only 9 samples. This imbalance makes evaluation challenging because models must learn discriminative features for underrepresented disease stages from very sparse examples. Therefore, overall accuracy alone may not fully describe model behavior, as strong performance on majority classes may mask weaker performance on minority classes. Hence, macro F1-score and per-class results are particularly useful for interpreting performance on this dataset.

Each individual model performs well, achieving accuracy between 97.16\% and 98.47\% and macro F1-score between 97.79\% and 98.85\%. Among individual models, ViT-Large performs best (98.47\% accuracy, 98.85\% macro F1), followed by Swin Transformer (98.36\%, 98.75\%). These results suggest that both global and hierarchical attention mechanisms are effective in capturing disease-relevant structural patterns. Among convolutional models, ReXNet-150 (97.81\%, 98.38\%) surpasses ResNet-50 (97.48\%, 97.79\%), while ViT-Base (97.16\%, 98.15\%) shows relatively balanced predictions despite lower overall accuracy, as reflected in its macro F1-score.

Despite severe class imbalance, performance remains consistently high across individual models. As shown in Table~\ref{tab:perclass_f1}, all models achieve an F1-score of 1.00 for the \textit{Moderate Demented} class in this split. However, this class contains only 9 test samples, so the result should not be overinterpreted. Lower F1-scores are generally observed for the \textit{Very Mild Demented} and \textit{Mild Demented} categories, indicating the difficulty of distinguishing adjacent early disease stages.

\subsection{Results on Dataset~II}

Dataset~II is the most challenging setting because of the mismatch between a balanced training set and a highly imbalanced test set, where \textit{No Impairment} accounts for 640 samples and \textit{Moderate Impairment} only 12 out of 1{,}279 test samples. ViT-Large obtains the best performance (97.81\% accuracy, 98.30\% macro F1), followed by the Swin Transformer (97.19\%, 97.60\%). ReXNet-150 (95.70\%, 95.80\%) and ResNet-50 (94.06\%, 93.00\%) perform lower, suggesting greater sensitivity of these convolutional models to distribution mismatch in this setting.

ViT-Base shows the largest degradation, with 91.30\% accuracy and 86.00\% macro F1. The large gap between accuracy and macro F1 indicates uneven performance across classes under distribution mismatch. As shown in Table~\ref{tab:perclass_f1}, ViT-Base obtains an F1-score of 0.72 for the \textit{Moderate Impairment} class, while ViT-Large and Swin Transformer achieve 1.00 in this split. However, the \textit{Moderate Impairment} class contains only 12 test samples, so these class-level results should be interpreted cautiously. Lower F1-scores are also observed for the \textit{Very Mild Impairment} and \textit{Mild Impairment} categories for several models, highlighting the difficulty of distinguishing adjacent early stages under distribution mismatch.

\subsection{Results on Dataset~III}

Dataset~III contains augmented MRI images with a larger sample size and less severe class imbalance, resulting in a 5{,}098-sample standard test set. The larger sample size and reduced imbalance provide a more stable setting for training and evaluation. All individual models achieve very high accuracy, ranging from 98.35\% (ViT-Base) to 99.84\% (ViT-Large), and macro F1-scores ranging from 98.50\% to 99.86\%. The difference between the best and worst individual models is much smaller than in Dataset~II, suggesting that the larger sample size and less severe class imbalance are associated with smaller performance differences across architectures.

ViT-Large achieves the best results among individual models
(99.84\% accuracy, 99.86\% macro F1), followed closely by the
Swin Transformer (99.57\% accuracy, 99.60\% macro F1).
ReXNet-150 (99.41\%, 99.46\%) outperforms ResNet-50
(98.53\%, 98.64\%) and ViT-Base (98.35\%, 98.50\%),
suggesting that efficient convolutional models can perform strongly in the larger and less imbalanced augmented-data setting. As shown in Table~\ref{tab:perclass_f1}, class-level performance is near ceiling across all individual models. The lowest F1-scores generally occur for the \textit{Non Demented} and \textit{Very Mild Demented} categories, while the \textit{Mild Demented} and \textit{Moderate Demented} classes achieve consistently high F1-scores. The reduced performance variance indicates that model performance approaches saturation under augmented-data conditions and that architectural differences become less pronounced.

\subsection{Cross-Dataset Study}

The three datasets provide complementary evaluation settings for examining model behavior under class imbalance, distribution mismatch, and data augmentation. Dataset~II is the most challenging of the three and produces the largest performance decline for most models, with balanced training data evaluated against a highly imbalanced test distribution. The magnitude of this decline varies considerably across architectures. ViT-Base shows the largest decrease, dropping by 5.86 accuracy points from Dataset~I to Dataset~II, while ViT-Large decreases by only 0.66 points. This suggests that model capacity may contribute to more consistent performance under the evaluated distribution mismatch, although architectural and optimization differences may also influence the result.

\begin{figure}[H]
  \centering
  \includegraphics[width=\columnwidth]{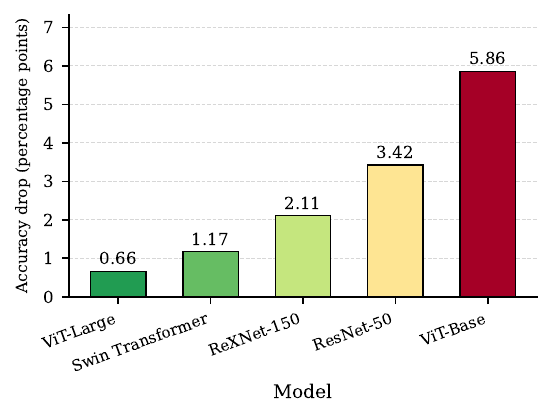}
  \caption{Accuracy decline from Dataset~I to Dataset~II for each model. Models are ordered by the magnitude of performance degradation under the evaluated train-test distribution mismatch. ViT-Large shows the smallest decrease (0.66 points), while ViT-Base shows the largest (5.86 points).}
  \label{fig:degradation}
\end{figure}

Performance differences between architectures become more pronounced under this challenging setting. The performance gap is relatively small on Datasets~I and~III but substantially larger on Dataset~II, indicating that architectures differ in their sensitivity to the evaluated distribution mismatch. Lower class-level performance is also generally observed for adjacent early-stage categories, particularly \textit{Very Mild} and \textit{Mild}, highlighting the difficulty of distinguishing subtle structural differences during early disease progression. This pattern indicates that performance consistency cannot be inferred from results on a single dataset alone, even when the same architectures achieve high absolute accuracy. The wider spread observed on Dataset~II also suggests that more challenging data conditions can reveal architectural differences that remain less visible in easier or more heavily augmented settings. Figure~\ref{fig:degradation} summarizes the architecture-specific accuracy decline from Dataset~I to Dataset~II.

\subsection{Summary of Experimental Results}

Model performance patterns remain broadly consistent across the three datasets, although the magnitude of performance differences varies with dataset conditions. ViT-Large and Swin Transformer consistently rank among the strongest individual models, while performance differences become more pronounced on Dataset~II under train-test distribution mismatch and less pronounced on the larger augmented Dataset~III.

Across the evaluated settings, lower class-level performance is observed mainly among adjacent early-stage categories, particularly \textit{Very Mild} and \textit{Mild} impairment. The near-ceiling performance on Dataset~III suggests that architectural differences become less apparent under augmented-data conditions, although these results may be influenced by redundancy or visually similar samples~\cite{yagis2021effect, tampu2022inflation, rouzrokh2022mitigating, truong2024benchmarking}.

Taken together, the findings indicate that both architectural characteristics and dataset conditions influence performance. Architecture differences become clearer under the more difficult distribution-mismatch setting and less apparent under the larger augmented-data setting. Independent clinical validation remains necessary to assess generalizability beyond the evaluated benchmark conditions.

\section{Limitations}
\label{sec:limitations}

The benchmark models achieve near-ceiling performance on Dataset~III. However, this result should be interpreted with caution because Dataset~III is based on augmented MRI images derived from a limited set of original samples. Visually similar or source-related images may appear across training, validation, and test splits, which can reduce evaluation difficulty and inflate performance. Therefore, Dataset~III results should be treated as augmented-data performance rather than evidence of clinical generalization.

Another limitation is that all datasets use similar four-class Alzheimer's disease staging labels and preprocessed structural MRI images. Therefore, the findings may not fully generalize to independent clinical cohorts, multi-site hospital data, or scans acquired under different imaging protocols. In addition, this study focuses only on image-level classification and does not include longitudinal information, clinical variables, cognitive scores, or demographic factors. Although preprocessing and evaluation are standardized, some optimization settings are adapted by dataset and architecture. Therefore, observed performance differences should be interpreted as comparative results under the selected fine-tuning configurations rather than as isolated causal effects of architecture or model capacity.

\section{Conclusions}
\label{sec:conclusion}

In this work, we present a multi-dataset benchmarking framework for Alzheimer's disease stage classification from MRI using five convolutional and transformer-based architectures across datasets with severe class imbalance, train-test distribution mismatch, and large-scale augmentation. ViT-Large achieves the strongest performance across all three datasets, while Swin Transformer consistently ranks among the top models. Performance differences are largest on Dataset~II and much smaller on the augmented Dataset~III, showing that dataset characteristics strongly influence the separation between architectures. Lower class-level performance is observed mainly among the Very Mild and Mild categories, highlighting the difficulty of distinguishing adjacent early disease stages. These findings show that evaluating models across multiple datasets provides a more reliable view of performance consistency than relying on a single dataset. Future work should include independent clinical cohorts, subject-level splits, multiple scanners and sites, and longitudinal or multimodal patient information.

\bibliographystyle{IEEEtran}
\bibliography{ref}

\makeatother

\end{document}